\documentclass{article}

\usepackage[nonatbib,preprint]{nips_2018}

\usepackage[utf8]{inputenc} 
\usepackage[T1]{fontenc}    
\usepackage{hyperref}       
\usepackage{url}            
\usepackage{booktabs}       
\usepackage{amsfonts}       
\usepackage{nicefrac}       
\usepackage{microtype}      
\usepackage{graphicx}
\usepackage{color}
\usepackage{multirow}

\title{An Empirical Analysis of Deep Audio-Visual Models for Speech Recognition}

\author{
  Devesh Walawalkar,\ \ Yihui He,\ \ Rohit Pillai \\
  Carnegie Mellon University\\
  Pittsburgh, PA 15213 \\
  \texttt{\{dwalawal, he2, rrpillai\}@andrew.cmu.edu} \\
}

\begin{document}

\maketitle

\begin{abstract}
In this project, we worked on speech recognition, specifically predicting individual words based on both the video frames and audio~\cite{chung2016lip,Chung17a,chung2017lip}. Empowered by convolutional neural networks~\cite{alexnet,vgg}, the recent speech recognition and lip reading models are comparable to human level performance~\cite{assael2016lipnet,torfi20173d}.  We re-implemented and made derivations of the state-of-the-art model presented in \cite{end2end}. Then, we conducted rich experiments including the effectiveness of attention mechanism~\cite{chan2016listen}, more accurate residual network~\cite{resnet} as the backbone with pre-trained weights and the sensitivity of our model with respect to audio input with/without noise. 

\end{abstract}

\section{Introduction}
In recent years, very deep convolutional neural networks (CNNs)~\cite{zf,xception,mnasnet,addressnet} have led to a series of breakthroughs in many audio and image understanding problems~\cite{imagenet,coco,zhou2017scene,krishna2017visual,lightweight}, such as image recognition~\cite{resnet,googlenet,cp,amc}, object detection~\cite{fast,faster,softer}, video surveillance~\cite{ma2018vehicle,wang2014hierarchical} and speech recognition~\cite{petridis2016prediction}. In particular, the recent speech recognition and lip reading models empowered by CNNs are comparable to human level performance~\cite{assael2016lipnet,torfi20173d,stafylakis2017combining,heimbach2018empirical,afouras2018deep,afouras2018conversation,sterpu2018attention,shillingford2018large}.

Audio-visual speech recognition systems usually consist of three parts~\cite{potamianos2003recent,dupont2000audio}: visual, audio and fusion. 
First, the visual part detects and tracks a speaker's lip movements and extracts relevant speech features. 
Second,  noise-robust features are extracted from the audio signals with the acoustic part.
Third, the fusion module is responsible for joint training of the audio-visual streams using models such as hidden Markov Models (HMM), deep networks with the gated recurrent unit (GRU).

In this project, we first re-implemented a deep end-to-end model for audio-visual recognition~\cite{end2end}. To the best of our knowledge, it is the state-of-the-art approach to tackling this problem. Specifically, we used a multi-modal deep learning model to learn the words pronounced in a particular time frame. We trained a CNN based model infused with temporal learning to extract temporal features from the video frames. We used a combination of 3D convolution and GRU techniques to learn the same.\\
We also trained a separate model for extracting the features from the audio present in the video. Pronunciation of the words is an essential factor which distinguishes words having the same lip movement, thus making classification of words based on only the visual cues very difficult. We used similar GRU techniques as for the video to extract the audio features.\\
We then combined both these models using Bidirectional GRU to learn features from the combination of the frames and audio. This is then given to a soft-max layer to predict the respective classes. The detailed model architecture is shown in Figure~\ref{fig:model}. To show the generality of our method, we tested on Lip Reading in the Wild (LRW) BBC dataset \cite{chung2016lip} that has up to 1000 utterances of the same word for more than 500 different words.

We summarize our contributions as follow: 
\begin{enumerate}
    \item We re-implemented and made derivations of the state-of-the-art model presented in \cite{end2end}.
    \item We introduced attention mechanism~\cite{chan2016listen} to our model, which improves the performance by around 4\% for the video only model, by around 2\% for the audio-only model and by around 1\% for the combined model.
    \item We replaced the ResNet model~\cite{resnet} with a more accurate CNN model with pre-trained weights, which interestingly improves the performance. 
    \item We further studied the sensitivity of our model with respect to audio input with/without noise.
\end{enumerate}

\section{Related works}
Earlier solutions to speech recognition mostly used either classical signal processing techniques or deep learning on only the video data or audio data to do the actual recognition. In the video space, LipNet \cite{assael2016lipnet} is one example where a CNN is used with bi-directional GRU's to predict the word being said in the current frame using the sequence of words said before. It then uses these frame wise predictions to determine the optimal sequence of predicted words. Similarly, Chung et al. \cite{chung2016lip} built multiple CNNs based on the architecture of VGG-M that would use 25 fps to detect words from a sequence of lip movements. \cite{stafylakis2017combining} also uses spatiotemporal convolutions to generate a prediction for the word being said in the current frame after landmarking and using standard 3D convolutions to augment the input video data.

From \cite{SpeechRecognitionReview}, we see that speech recognition has evolved from the classical techniques of phoneme matching, which assumed that all sounds could be produced from a fixed number of sounds to pattern matching that is based on a solid mathematical background. Pattern matching involved learning the structure of audio waveforms during training using an HMM, a template or some other construct and then comparing these learned structures with the test input to match the best one. These techniques gave way to the state of the art knowledge-based approaches that use machine learning. One of the earliest knowledge-based approaches was the SVM, that was severely limited in the fact that it could not be used to translate variable length sentences into text. Every test input had to convert to a fixed size sentence before the SVM could classify it. However, the concept of feature extraction led to the development of several techniques to accurately extract the essential features to a sound (using PCA, LDA, ICA, kernel-based feature extraction, etc.) that would then be fed into a classifier to produce an output. Nowadays, neural networks have replaced all these techniques since they are able to learn and extract much more complex features than any of the previously mentioned techniques were. In terms of state of the art techniques, \cite{Graves2014TowardsES} published by DeepMind describes a network architecture that transcribes speech to text without intermediary representations of this data. The architecture involves bidirectional LSTM's as the hidden layers in a bidirectional deep RNN. In addition to this, the objective function used to train the network is the Connectionist Temporal Classification function. 

Deep architectures that use both audio and video data also tend to use LSTM or GRU  units for their predictions. This is seen in the encoder-decoder architecture employed in \cite{chung2017lip} which uses unidirectional LSTMs to encode both the image and audio data and generates attention vectors to predict the word being said.  While \cite{stafylakis2017combining} used only video frames as input, it can be easily extended to incorporate both audio and visual information as seen in \cite{end2end}. This uses two separate ResNets and BGRUs to extract features and model temporal dependencies from the visual and audio inputs and two additional BGRU's to combine the extracted audio and visual features. \cite{7780758} uses another approach which uses temporal multimodal networks to learn a joint distribution over a mouth and lip movements along with the audio at every frame. These joint distributions are then combined to get a time-dependent sequence of frames and audio. 

\begin{figure}[t]
\begin{center}
    \includegraphics[scale=0.35]{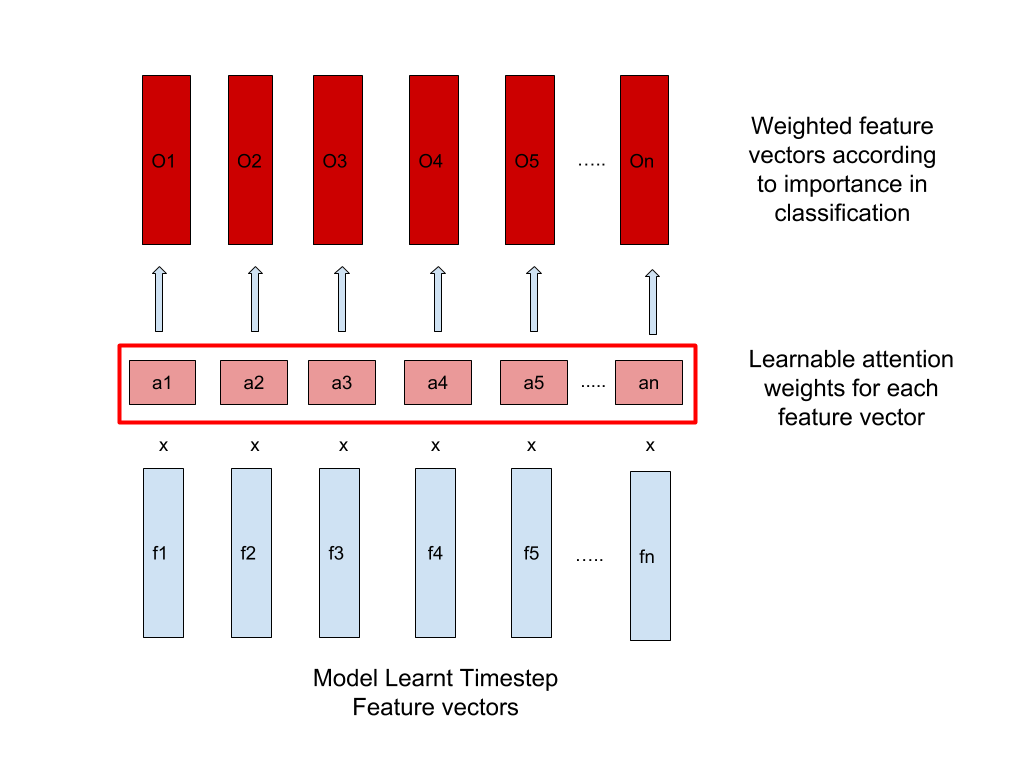}
\end{center}
    \caption{ Implemented Attention mechanism}
    \label{fig:model}
\end{figure}

\section{Dataset}

The dataset used for this project was the LRW BBC dataset \cite{chung2016lip} that has up to 1000 utterances of the same word for more than 500 different words. Each utterance is taken from different BBC presenters and guests on the air and is presented in the form of a video that is 29 frames (1.16 seconds) long. Every video also has metadata associated with it that indicates the duration of the word, which allows us to determine the start and end frames of the word being uttered. This dataset is challenging since there may be multiple words being said in a video, and as a result, the lip movements may be influenced by both past and future words in addition to the word we are trying to learn. Some of the classes are also very similar to one another (different tenses of the same word, singular vs. plural) which makes this dataset even more challenging. In total, there are 538786 different videos out of which 488786 are training examples, 25000 were for validation, and 25000 were used to test. 

\section{Audio Visual Recognition}

\subsection{Attention Mechanism}\label{section:attention}

Not all frames and their corresponding audio snippets are equally important in learning to classify the word. In order to take this into account, we introduce the concept of attention into our model. For our model, attention is just a 1-D vector that is the same size as the output of the previous layer's time-step count, and each value in the vector is multiplied across all features of a particular time-step. These values, which range from 0 to 1, can be learned and help us extract the more relevant features from our inputs.
For the final model output of size $[time steps,Features]$ we multiply all features of a single time-step feature vector $[1,Features]$ with a single element from the $[time steps,]$ dimensional vector learned by the model. In our experiments, we found out the vector learns a type of Gaussian distribution across the length of the vector, i.e. the centre elements have values in the range of 0.8 to 0.95, while those at the end and the beginning have values in the range of 0.1 to 0.2. This follows intuitively that the frames/audio slices at the middle are more important for classification compared to the ones at the end and the beginning.\\
We incorporate three types of attention in our models. They are as follows:
\begin{enumerate}
    \item  \textbf{Video attention}: This attention is used to weigh the importance of time-step feature vectors outputted by the video sub model alone.
    \item \textbf{Audio attention}: This attention is used to weigh the importance of time-step feature vectors outputted by the audio sub model alone.
    \item  \textbf{Combined attention}: This attention is used to weigh the importance of time-step feature vectors outputted by the combined sub model at the final stages of the overall model. The combined attention weighted output is directly fed to the classification layer.
\end{enumerate}

\subsection{Visual sub model}
The visual model consists of a spatiotemporal convolution followed by a 34-layer ResNet and a 2-layer BGRU. A spatiotemporal convolutional layer is capable of capturing the short-term dynamics of the mouth region. It consists of a convolutional layer with 3D kernels of 5 by 7 by 7 size (time/width/height), followed by batch normalization and rectified linear units. Once this convolution is done, we then feed it through a Resnet 34 that reduces the dimensionality of the input such that it outputs a 1-D tensor. This tensor is then passed through a two-layer BGRU's of 1024 units each. Finally, we multiply it with a visual attention vector where each frame has a single attention value associated with it.

\subsection{Audio sub model}
The audio model consists of an 18-layer ResNet followed by two BGRU layers. We use the standard architecture for the ResNet-18 with the main difference being that we use 1D instead of 2D kernels which are used for frame data. The output of the ResNet is divided into 29 frames/windows to ensure that there is a 1-1 correspondence between a video frame and an audio snippet (29 for each training example)
The output of the ResNet-18 is fed to a 2-layer Bi-GRU which consists of 1024 cells in each layer. The output from the second Bi-GRU is also multiplied by an attention vector of length 29 with each element corresponding to one snippet of the audio.

\subsection{Overall model}
Shown in Figure~\ref{fig:model}, for the overall combined model, outputs of each sub-model are concatenated and fed to
another 2-layer BGRU of 1024 units each in order to fuse the information from the audio and visual streams and jointly model their temporal dynamics~\cite{end2end}. The output of the 2-dimensional BGRU is then multiplied by another attention vector which is also of length 29. Finally, the output layer is a softmax layer which provides a label to each frame. The sequence is labeled based on the highest average probability. 

\begin{figure}[t]
\begin{center}
\includegraphics[scale=0.35]{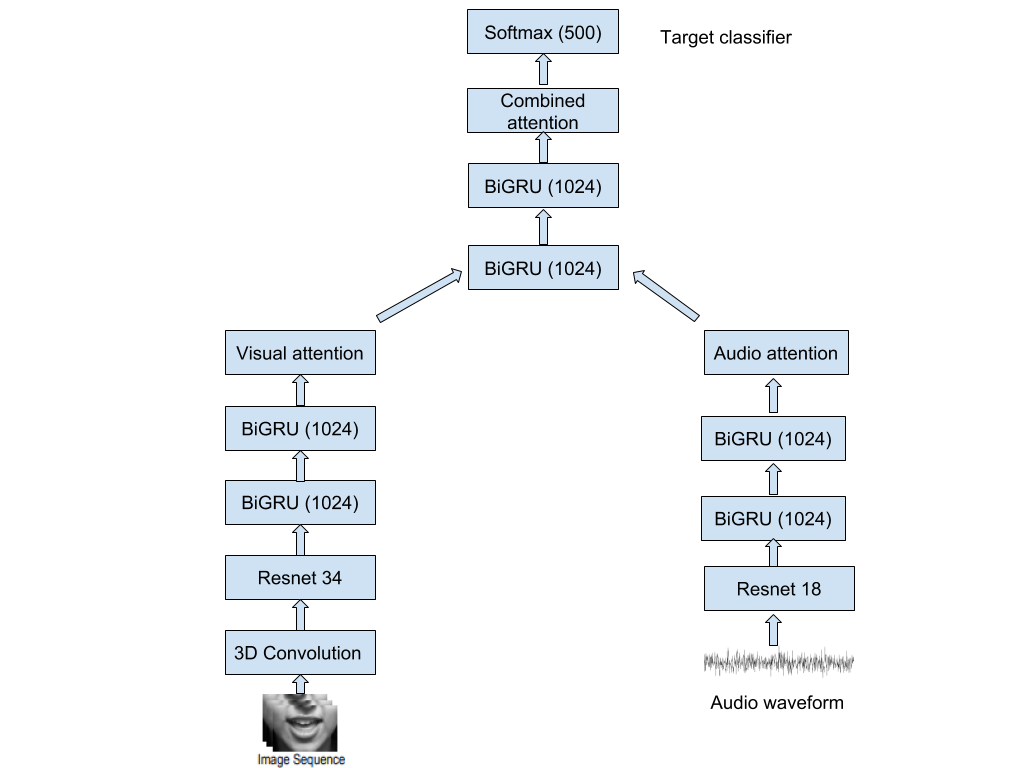}
\end{center}
    \caption{Combined audio-visual model.}
    \label{fig:model}
\end{figure}

\section{Experimental setup}
Our implementation is based on PyTorch~\cite{pytorch}.

We trained the audio model end-to-end using only the audio data. Similarly, we trained the visual model end to end using only the frames extracted. We then plan to combine both the trained models and train the entire system end to end using both the types of data available. During the last stage of training, we plan to keep the weights of the base learned layers fixed, while only fine-tuning on the front combining Bi-GRU network.  

\subsection{Dataset Pre-processing}
For our model training, we are using the Lip Reading Words (LRW) dataset. This dataset consists of up to 1000 utterances of 500 different words spoken during a short video clip. For the purpose of our model, we have extracted the frames and audio from these clips separately.
We extracted the frames from each video example at the frame rate of 29 FPS. We then extract a region of the fixed area around the mouth for every sample. All video samples in the dataset have the mouth located in a specific region, which made it easy for us to extract it.
We extracted the audio from the videos at the rate of 16KHz. We converted the audio files into numpy arrays so as to make it possible to feed this extracted information to our audio sub-model.  

\subsection{Training}

Training for our model is done in 2 stages. We first learn the weights for our individual streams by training them on the audio and video streams separately. Once we have the weights for these layers, we then train the entire combined model end to end. While training, we also add Gaussian noise to both our video and audio inputs so that our network is more robust to different inputs. We just add Gaussian noise to the audio signal while for the video, we flip a frame or randomly crop it with probability 0.5.

\subsubsection{Training Individual Streams}

Since training each individual stream end-to-end is sub-optimal, the stream is trained using a 3 step procedure. For the first step, the 2 layers Bi-GRU is replaced by a temporal convolutional back end. The entire system (the ResNet, the temporal convolutional back end, the attention layer, and a softmax) is then trained until there is no improvement in accuracy for more than five epochs. Once this is achieved, the 2 layers Bi-GRU back is inserted and trained for five epochs with all the other weights (ResNet and attention layers) kept constant. \\
Once the ResNet's and two-layer Bi-GRU's weights are computed, they are put together and trained end to end with a softmax layer as the output. The system was trained with the Adam optimizer algorithm with learning rate = 0.0001 for the whole system except the attention layer whose learning rate was 0.0002 and batch size = 32.

\subsubsection{End-to-end model training}

The weights from the trained single streams were used to initialize the corresponding components in the final architecture. The outputs of the two streams were fed into another 2 layers Bi-GRU followed by an attention layer. The additional layers were trained for five epochs with all the weights in the individual streams not changed. After this, the entire network is trained with the Adam optimizer with an initial learning rate of 0.0001 for everything except the attention layers which had 0.0002 as the initial learning rate. We incorporated stepwise learning rate decay after every ten epochs, with a batch size of 32.

\section{Results}
As mentioned before, we make three modifications for our deep learning model experiments. Following are the results obtained: 

\subsection{Using attention mechanism}
We introduce the concept of attention into our model in section~\ref{section:attention} since not all frames and their corresponding audio snippets are equally important in learning to classify the word. 
 Surprisingly using the attention mechanism~\cite{chan2016listen}, the performance is improved by around 4\% for the video only model, by around 2\% for the audio only model and by around 1\% for the combined model, shown in table~\ref{table:attention mechanism}. We observed that the improvement for the combined model is limited (1\%), since the baseline accuracy is already quite high (97.43\%).
\begin{table}[h]
\begin{center}
\begin{tabular}{c|c|c}
\hline
& Without attention & With attention \\ \hline
Audio & 0.9594 & 0.9702 \\ \hline
Visual & 0.8290 & 0.8617   \\ \hline
AudioVisual & 0.9743  & 0.9823  \\ \hline
\end{tabular}
\end{center}
\caption{Inclusion of attention mechanism}
\label{table:attention mechanism}
\end{table}

\subsection{The Effectiveness of Noise Input}
During training, we observed that the training accuracy could easily reach 99\%, even 100\% since the dataset is not challenging enough for deep neural networks. This motivates us to analyze the effectiveness of noise input.
Shown in Table~\ref{table:noise addition}, we studied the effectiveness of noise input. With noise input, the accuracy is consistently improved for audio, visual and audiovisual models (0.9\%, 0.25\%, 0.41\% respectively). 
\begin{table}[h]
\begin{center}
\begin{tabular}{c|c|c}
\hline
& Without noise & With noise \\ \hline
Audio & 0.9702 & 0.9792 \\ \hline
Visual & 0.8617 & 0.8642   \\ \hline
AudioVisual & 0.9823  & 0.9864  \\ \hline
\end{tabular}
\end{center}
\caption{The Effectiveness of Noise Input}
\label{table:noise addition}
\end{table}

\subsection{Making DNN model deeper}
Deep residual learning~\cite{resnet} found that the performance is usually better when the convolutional neural network is deeper for the image classification task. This motivates us to verify this conclusion on our speed recognition and lip reading task.
Shown in Table~\ref{table:Deeper model}, we studied the effectiveness of deeper DNN model, namely ResNet-34. With ResNet-34, the accuracy is consistently improved across all three models (0.18\%, 0.07\%, and 0.21\% improvement for audio, visual and audiovisual models respectively). We observed that the improvement is marginal and the improvement for audio model is slightly larger. 
\begin{table}[h]
\begin{center}
\begin{tabular}{c|c|c}
\hline
& With ResNet 18 & With ResNet 34 \\ \hline
Audio & 0.9702 & 0.9720 \\ \hline
Visual & 0.8617 & 0.8624   \\ \hline
Audio-Visual & 0.9823  & 0.9842  \\ \hline
\end{tabular}
\end{center}
\caption{Making model deeper}
\label{table:Deeper model}
\end{table}

\subsection{Overall comparison}
Finally, we combined our modifications and demonstrated the overall comparison shown in Table~\ref{table:Overall comparison}. Empirically, with attention mechanism and noise input, our audio-visual combined model achieves 98.64\%.
\begin{table}[h]
\begin{center}
\begin{tabular}{c|c|c|c|c|c}
\hline
  & \multicolumn{3}{c|}{w/o adding noise to data} & \multicolumn{2}{c}{Adding noise to data} \\
Model & \multicolumn{3}{c|}{} & \multicolumn{2}{c}{} \\\cline{2-6}
Type & Petridis et al. & Ours & Ours & Petridis et al.  & Ours   \\
& \cite{end2end} & [w/o Attention] &  [w/ Attention] & \cite{end2end} & [w/ attention] \\ \hline 
Video only & 0.8246  & 0.8290 & 0.8617 & 0.8300 & 0.8642  \\ \hline
Audio only & 0.9578  & 0.9594 & 0.9702 & 0.9717 & 0.9792  \\ \hline
AudioVisual & 0.9720  & 0.9743 & 0.9823 & 0.9800 & 0.9864  \\ \hline

\end{tabular}
\end{center}
\caption{Overall comparison}
\label{table:Overall comparison}
\end{table}

\section{Conclusion}
We re-implemented and made derivations of the model presented in \cite{end2end}. We proposed a novel attention mechanism for the model and obtained improved state-of-the-art results on it. We replaced the ResNet model with a more accurate CNN model with pre-trained weights. We also studied the sensitivity of our model with respect to audio input with/without noise and found considerable accuracy gain by incorporating it in our model.

\section{Future Works}
We plan on training the complete model end-to-end with focal loss~\cite{fl} to obtain better results. Besides, on the data processing part, we intend to sample the frames and audio at much higher sampling rates to extract much richer features to train the model. 

\bibliography{nips_2018}

\begin{thebibliography}{10}

\bibitem{afouras2018conversation}
T.~Afouras, J.~S. Chung, and A.~Zisserman.
\newblock The conversation: Deep audio-visual speech enhancement.
\newblock {\em arXiv preprint arXiv:1804.04121}, 2018.

\bibitem{afouras2018deep}
T.~Afouras, J.~S. Chung, and A.~Zisserman.
\newblock Deep lip reading: a comparison of models and an online application.
\newblock {\em arXiv preprint arXiv:1806.06053}, 2018.

\bibitem{assael2016lipnet}
Y.~M. Assael, B.~Shillingford, S.~Whiteson, and N.~de~Freitas.
\newblock Lipnet: End-to-end sentence-level lipreading.
\newblock {\em arXiv preprint arXiv:1611.01599}, 2016.

\bibitem{chan2016listen}
W.~Chan, N.~Jaitly, Q.~Le, and O.~Vinyals.
\newblock Listen, attend and spell: A neural network for large vocabulary
  conversational speech recognition.
\newblock In {\em Acoustics, Speech and Signal Processing (ICASSP), 2016 IEEE
  International Conference on}, pages 4960--4964. IEEE, 2016.

\bibitem{xception}
F.~Chollet.
\newblock Xception: Deep learning with depthwise separable convolutions.
\newblock {\em arXiv preprint}, pages 1610--02357, 2017.

\bibitem{chung2017lip}
J.~S. Chung, A.~W. Senior, O.~Vinyals, and A.~Zisserman.
\newblock Lip reading sentences in the wild.
\newblock In {\em CVPR}, pages 3444--3453, 2017.

\bibitem{chung2016lip}
J.~S. Chung and A.~Zisserman.
\newblock Lip reading in the wild.
\newblock In {\em Asian Conference on Computer Vision}, pages 87--103.
  Springer, 2016.

\bibitem{Chung17a}
J.~S. Chung and A.~Zisserman.
\newblock Lip reading in profile.
\newblock In {\em British Machine Vision Conference}, 2017.

\bibitem{dupont2000audio}
S.~Dupont and J.~Luettin.
\newblock Audio-visual speech modeling for continuous speech recognition.
\newblock {\em IEEE transactions on multimedia}, 2(3):141--151, 2000.

\bibitem{fast}
R.~Girshick.
\newblock Fast r-cnn.
\newblock In {\em Proceedings of the IEEE International Conference on Computer
  Vision}, pages 1440--1448, 2015.

\bibitem{Graves2014TowardsES}
A.~Graves and N.~Jaitly.
\newblock Towards end-to-end speech recognition with recurrent neural networks.
\newblock In {\em ICML}, 2014.

\bibitem{resnet}
K.~He, X.~Zhang, S.~Ren, and J.~Sun.
\newblock Deep residual learning for image recognition.
\newblock In {\em Proceedings of the IEEE conference on computer vision and
  pattern recognition}, pages 770--778, 2016.

\bibitem{amc}
Y.~He, J.~Lin, Z.~Liu, H.~Wang, L.-J. Li, and S.~Han.
\newblock Amc: Automl for model compression and acceleration on mobile devices.
\newblock In {\em Proceedings of the European Conference on Computer Vision
  (ECCV)}, pages 784--800, 2018.

\bibitem{addressnet}
Y.~He, X.~Liu, H.~Zhong, and Y.~Ma.
\newblock Addressnet: Shift-based primitives for efficient convolutional neural
  networks.
\newblock In {\em 2018 IEEE Winter Conference on Applications of Computer
  Vision (WACV)}. IEEE, 2019.

\bibitem{softer}
Y.~He, X.~Zhang, M.~Savvides, and K.~Kitani.
\newblock Softer-nms: Rethinking bounding box regression for accurate object
  detection.
\newblock {\em arXiv preprint arXiv:1809.08545}, 2018.

\bibitem{cp}
Y.~He, X.~Zhang, and J.~Sun.
\newblock Channel pruning for accelerating very deep neural networks.
\newblock In {\em International Conference on Computer Vision (ICCV)},
  volume~2, 2017.

\bibitem{heimbach2018empirical}
K.~Heimbach.
\newblock An empirical evaluation of convolutional and recurrent neural
  networks for lip reading.
\newblock Master's thesis, 2018.

\bibitem{7780758}
D.~Hu, X.~Li, and X.~Lu.
\newblock Temporal multimodal learning in audiovisual speech recognition.
\newblock In {\em 2016 IEEE Conference on Computer Vision and Pattern
  Recognition (CVPR)}, pages 3574--3582, June 2016.

\bibitem{krishna2017visual}
R.~Krishna, Y.~Zhu, O.~Groth, J.~Johnson, K.~Hata, J.~Kravitz, S.~Chen,
  Y.~Kalantidis, L.-J. Li, D.~A. Shamma, et~al.
\newblock Visual genome: Connecting language and vision using crowdsourced
  dense image annotations.
\newblock {\em International Journal of Computer Vision}, 123(1):32--73, 2017.

\bibitem{alexnet}
A.~Krizhevsky, I.~Sutskever, and G.~E. Hinton.
\newblock Imagenet classification with deep convolutional neural networks.
\newblock In {\em Advances in neural information processing systems}, pages
  1097--1105, 2012.

\bibitem{imagenet}
A.~Krizhevsky, I.~Sutskever, and G.~E. Hinton.
\newblock Imagenet classification with deep convolutional neural networks.
\newblock In {\em Advances in neural information processing systems}, pages
  1097--1105, 2012.

\bibitem{lightweight}
Y.~Liang, Z.~Yang, K.~Zhang, Y.~He, J.~Wang, and N.~Zheng.
\newblock Single image super-resolution via a lightweight residual
  convolutional neural network.
\newblock {\em arXiv preprint arXiv:1703.08173}, 2017.

\bibitem{fl}
T.-Y. Lin, P.~Goyal, R.~Girshick, K.~He, and P.~Doll{\'a}r.
\newblock Focal loss for dense object detection.
\newblock {\em IEEE transactions on pattern analysis and machine intelligence},
  2018.

\bibitem{coco}
T.-Y. Lin, M.~Maire, S.~Belongie, J.~Hays, P.~Perona, D.~Ramanan,
  P.~Doll{\'a}r, and C.~L. Zitnick.
\newblock Microsoft coco: Common objects in context.
\newblock In {\em European conference on computer vision}, pages 740--755.
  Springer, 2014.

\bibitem{ma2018vehicle}
X.~Ma, Y.~He, X.~Luo, J.~Li, M.~Zhao, B.~An, and X.~Guan.
\newblock Vehicle traffic driven camera placement for better metropolis
  security surveillance.
\newblock {\em IEEE Intelligent Systems}, 2018.

\bibitem{pytorch}
A.~Paszke, S.~Gross, S.~Chintala, G.~Chanan, E.~Yang, Z.~DeVito, Z.~Lin,
  A.~Desmaison, L.~Antiga, and A.~Lerer.
\newblock Automatic differentiation in pytorch.
\newblock 2017.

\bibitem{petridis2016prediction}
S.~Petridis and M.~Pantic.
\newblock Prediction-based audiovisual fusion for classification of
  non-linguistic vocalisations.
\newblock {\em IEEE Transactions on Affective Computing}, 7(1):45--58, 2016.

\bibitem{end2end}
S.~Petridis, T.~Stafylakis, P.~Ma, F.~Cai, G.~Tzimiropoulos, and M.~Pantic.
\newblock End-to-end audiovisual speech recognition.
\newblock {\em arXiv preprint arXiv:1802.06424}, 2018.

\bibitem{potamianos2003recent}
G.~Potamianos, C.~Neti, G.~Gravier, A.~Garg, and A.~W. Senior.
\newblock Recent advances in the automatic recognition of audiovisual speech.
\newblock {\em Proceedings of the IEEE}, 91(9):1306--1326, 2003.

\bibitem{SpeechRecognitionReview}
D.~R. Reddy.
\newblock Speech recognition by machine: A review.
\newblock {\em Proceedings of the IEEE}, 64(4):501--531, April 1976.

\bibitem{faster}
S.~Ren, K.~He, R.~Girshick, and J.~Sun.
\newblock Faster r-cnn: Towards real-time object detection with region proposal
  networks.
\newblock In {\em Advances in neural information processing systems}, pages
  91--99, 2015.

\bibitem{shillingford2018large}
B.~Shillingford, Y.~Assael, M.~W. Hoffman, T.~Paine, C.~Hughes, U.~Prabhu,
  H.~Liao, H.~Sak, K.~Rao, L.~Bennett, et~al.
\newblock Large-scale visual speech recognition.
\newblock {\em arXiv preprint arXiv:1807.05162}, 2018.

\bibitem{vgg}
K.~Simonyan and A.~Zisserman.
\newblock Very deep convolutional networks for large-scale image recognition.
\newblock {\em arXiv preprint arXiv:1409.1556}, 2014.

\bibitem{stafylakis2017combining}
T.~Stafylakis and G.~Tzimiropoulos.
\newblock Combining residual networks with lstms for lipreading.
\newblock {\em arXiv preprint arXiv:1703.04105}, 2017.

\bibitem{sterpu2018attention}
G.~Sterpu, C.~Saam, and N.~Harte.
\newblock Attention-based audio-visual fusion for robust automatic speech
  recognition.
\newblock In {\em Proceedings of the 2018 on International Conference on
  Multimodal Interaction}, pages 111--115. ACM, 2018.

\bibitem{googlenet}
C.~Szegedy, W.~Liu, Y.~Jia, P.~Sermanet, S.~Reed, D.~Anguelov, D.~Erhan,
  V.~Vanhoucke, and A.~Rabinovich.
\newblock Going deeper with convolutions.
\newblock In {\em Proceedings of the IEEE conference on computer vision and
  pattern recognition}, pages 1--9, 2015.

\bibitem{mnasnet}
M.~Tan, B.~Chen, R.~Pang, V.~Vasudevan, and Q.~V. Le.
\newblock Mnasnet: Platform-aware neural architecture search for mobile.
\newblock {\em arXiv preprint arXiv:1807.11626}, 2018.

\bibitem{torfi20173d}
A.~Torfi, S.~M. Iranmanesh, N.~Nasrabadi, and J.~Dawson.
\newblock 3d convolutional neural networks for cross audio-visual matching
  recognition.
\newblock {\em IEEE Access}, 5:22081--22091, 2017.

\bibitem{wang2014hierarchical}
X.~Wang and Q.~Ji.
\newblock A hierarchical context model for event recognition in surveillance
  video.
\newblock In {\em Proceedings of the IEEE Conference on Computer Vision and
  Pattern Recognition}, pages 2561--2568, 2014.

\bibitem{zf}
M.~D. Zeiler and R.~Fergus.
\newblock Visualizing and understanding convolutional networks.
\newblock In {\em European conference on computer vision}, pages 818--833.
  Springer, 2014.

\bibitem{zhou2017scene}
B.~Zhou, H.~Zhao, X.~Puig, S.~Fidler, A.~Barriuso, and A.~Torralba.
\newblock Scene parsing through ade20k dataset.
\newblock In {\em Proceedings of the IEEE Conference on Computer Vision and
  Pattern Recognition}, 2017.

\end{thebibliography}
\bibliographystyle{abbrv}

\end{document}